\documentclass[11pt,a4paper]{article}
\usepackage[hyperref]{acl2020}
\usepackage{times}
\usepackage{latexsym}
\usepackage{amsmath}
\usepackage{amsfonts}
\usepackage{booktabs}
\usepackage{physics}
\usepackage{amssymb}

\usepackage{graphicx}

\usepackage{microtype}

\aclfinalcopy % Uncomment this line for the final submission
\newcommand{\Tabref}[1]{Table~\ref{#1}}

\newcommand{\appref}[1]{Appendix~\ref{#1}}
\title{On Explaining Your Explanations of BERT: \\An Empirical Study with Sequence Classification}

\author{Zhengxuan Wu \\
  Stanford University \\
  \texttt{wuzhengx@stanford.edu} \\\And
  Desmond C. Ong  \\
  National University of Singapore \\
  \texttt{dco@comp.nus.edu.sg}
}

\date{}

\begin{document}
\maketitle
\begin{abstract}
BERT, as one of the pretrianed language models, attracts the most attention in recent years for creating new benchmarks across GLUE tasks via fine-tuning. One pressing issue is to open up the blackbox and explain the decision makings of BERT. A number of attribution techniques have been proposed to explain BERT models, but are often limited to sequence to sequence tasks. In this paper, we adapt existing attribution methods on explaining decision makings of BERT in sequence classification tasks. We conduct extensive analyses of four existing attribution methods by applying them to four different datasets in sentiment analysis. We compare the reliability and robustness of each method via various ablation studies. Furthermore, we test whether attribution methods explain generalized semantics across semantically similar tasks. Our work provides solid guidance for using attribution methods to explain decision makings of BERT for downstream classification tasks.
\end{abstract}

\section{Introduction}
BERT, as one of the pretrained masked language models, can be fine-tuned to outperform many existing benchmarks in NLP community~\cite{devlin2019bert}. Fine-tuning with pretrained BERT often becomes the the \emph{de facto} go-to way for establishing benchmarks in NLP. Additionally, more advanced BERT-variants have been developed since the debut of BERT, such as RoBERTa~\cite{liu2019roberta}, ELECTRA~\cite{clark2020electra} and XLNet~\cite{yang2019xlnet}. 

While creating new benchmarks remains the most practical problem to solve, explaining weights learnt by powerful models such as BERT becomes another pressing issue. More importantly, understanding these models increases transparency of deep neural networks. This further benefits in solving real-world problems such as feature importance analyses for diseases diagnoses with medical images~\cite{yang2018explaining, bohle2019layer}.

% More importantly, being able to explain how these attention-based models make their decisions increases transparency of deep learning models.

% Recent efforts to increase transparency of self-attention based networks suggest that local attention weights within heads encode syntactic information~\cite{tenney2019bert} such as anaphora~\cite{voita2018context, goldberg2019assessing}, Parts-of-Speech~\cite{vig2019analyzing} and dependencies~\cite{raganato2018analysis, hewitt2019structural, clark2019what}. Researchers also implement gradient-based LRP to analyzing relevance of each head within the Transformer~\cite{}. Building on these studies, we seek to determine what features in the input layer are important for the model to make the the final top-layer prediction. 

Recently, a variety of efforts have been tried to explain BERT. Previous works suggested that local attention weights within heads encode syntactic information~\cite{tenney2019bert} such as anaphora~\cite{voita2018context, goldberg2019assessing}, Parts-of-Speech~\cite{vig2019analyzing} and dependencies~\cite{raganato2018analysis, hewitt2019structural, clark2019what}. In parallel, various attribution methods were adapted to explain BERT, such as analyzing head importance~\cite{voita-etal-2019-analyzing}, and probing structural properties learnt for sequence-to-sequence tasks~\cite{hao2020self}. A number of studies also proposed methods to explain self-attention models via learnt attention weights directly~\cite{wu-etal-2020-structured, abnar-zuidema-2020-quantifying}.

% explaining model decision makings against input features  understand decision makings of BERT models such as gradient sensitivity analysis~\cite{}, layerwise relevance propagation~\cite{} and layerwise attention tracing~\cite{}. These methods demonstrate that BERT is able to pick out features that are important as human readers~\cite{}, or weight heads inside BERT differently according to their attributed relevance~\cite{}.

However, a majority of previous works focus on explaining BERT with sequence-to-sequence tasks~\cite{voita-etal-2019-analyzing}. The validity of these studies is hard to justify due to the lack of verifiable ground truth. Here, we aim to close the gap by investigating the validity of different attribution methods through the lens of sequence classification tasks in sentiment analysis. One benefits of this approach is its face validity, where human has strong intuitions about semantics in sentiment understandings.

% for sequence classification tasks. One benefits of studying attribution methods for classification task is its face validity, where human has strong intuitions about semantics. Here, we aim to close the gap by investigating the validity of different attribution methods through the lens of sequence classification tasks in sentiment analysis. 

In this paper, we first introduce four widely used attribution methods, and adapt them to BERT with a classification head. Then, we apply these methods to analyze BERT models trained with four sentiment analysis tasks with similar semantics. Our contributions are two-fold: first, we study the validity and robustness of four widely used attribution methods with BERT models in classification tasks. Second, we provide extensive evidences on whether these attribution methods produce generalizable explanations over semantics across tasks.

\section{Attribution Methods}~\label{sec:method}
Given our neural classifier (i.e., BERT with a classification head) parameterized by $f_{c}(x)$ to predict probability of a input sequence being class $c$, attribution methods produce a relevance score $\mathbf{R} (x)$ of a token $x$ denotes the relevancy of this token w.r.t. our class of interest $c$~\footnote{To derive $\mathbf{R} (x)$, we use L2 norms for gradient-based methods, and absolute sum for LRP.}. Furthermore, $\mathbf{R}_{i} (x)$ represents the relevancy of $i$-th dimension of the token embedding. In other words, attribution methods can quantify whether a feature of a token is important in our classifier's decisions to predict class $c$. Note that $f_{c}(x)$ often only contains non-zero entries for the class $c$, which is either \emph{positive} for binary classification or \emph{very positive} for five class classification~\footnote{Code implementation is released at ~\url{https://github.com/frankaging/BERT_LRP}}.

\subsection{Gradient Sensitivity}
Gradient-based attribution methods such as Gradient Sensitivity (GS) relies on gradients over inputs~\cite{li2016visualizing}:
\begin{align}
    \mathbf{R}^{\text{GS}}_{i} (x) =  \frac{\partial f_{c}(x)}{\partial x_{i}} \label{eqn:uni-gradient}
\end{align}
where the left-hand side represents the derivative of the output w.r.t. a $i$-th dimension of $x$.

\subsection{Gradient $\times$ Input}
Building on top of GS, gradient $\times$ Input (GI) adds a element-wise product in Eqn.~\ref{eqn:uni-gradient} with $x_{i}$~\cite{kindermans2017reliability}:
\begin{align}
    \mathbf{R}^{\text{GI}}_{i} (x) = x_{i} \cdot \mathbf{R}^{\text{GS}}_{i} (x) \label{eqn:uni-gradient-input}
\end{align}
Intuitively, gradients measure how significantly the output will change when a feature is perturbed.

\subsection{Layerwise Relevance Propagation}
To derive Layerwise Relevance Propagation (LRP) for BERT, we start with the simplest case where the neural network contains only linear layers with non-linear activation functions $g(\cdot)$:
\begin{align}
    z_{ij}^{l+1} = x_{i}^{l} w_{ij}^{l}\quad & z_{j}^{l+1} = \sum_{i} z_{ij}^{l+1} + b_{i} \label{eqn:linear-layer-act-1} \\
    x_{j}^{l+1} &= g(z_{j}^{l+1})
    \label{eqn:linear-layer-act-2}
\end{align}
where $w_{ij}$ is the weight edge connection neurons between layers $i$ and $j$ where $j > i$. LRP for such network can then be derived as: 
\begin{align}
    \mathbf{R}^{\text{LRP}}_{i} (x) &= f_{c}(x) ( \frac{\mathbf{w}^{l} x^{l}}{\mathbf{z}^{l+1}}) g^{\prime}(z_{j}^{l+1}) \\
    &\quad \cdots (\frac{\mathbf{w}^{0} x}{\mathbf{z}^{1}}) g^{\prime}(z_{j}^{1}) \label{eqn:lrp-chain-rule-1} \\
    &= f_{c}(x) (\prod_{l} \frac{\mathbf{w}^{l} x^{l}}{\mathbf{z}^{l+1}}) (\prod_{l} g^{\prime}(z_{j}^{l+1})) \label{eqn:lrp-general-1} \\
    % &= f_{c}(x) (\prod_{l} \frac{z^{l}}{\mathbf{z}^{l+1}}) (\prod_{l} g^{\prime}(z_{j}^{l+1})) \label{eqn:lrp-general-3} \\
    &\approx f_{c}(x) (\prod_{l} \frac{z^{l}}{\mathbf{z}^{l+1}})
\end{align}
where $\mathbf{z}^{l}$ is column matrix of hidden states in layer $l$, and derivatives of non-linear activation functions $g^{\prime}(\cdot)$ are ignored as proposed in~\citet{bach2015pixel}. See Appendix~\ref{app:non-linear} for justifications.

To deduct full LRP for BERT, we need to derive $z^{l}$ for non-linear layers such as the self-attention layer and the residual layer. we use the first term in the Taylor expansion to approximate $z^{l}$ as proved in~\cite{bach2015pixel}:
\begin{align}
    z^{l} &\approx \frac{\partial f_{\psi}({x_{i}^{l-1}}) }{\partial x_{i}^{l-1}} ( \mathbf{x}^{l-1} -  \mathbf{x}^{l-1}_{0} ) 
    \label{eqn:taylor}
\end{align}
where $\hat{\mathbf{x}}^{l-1}$ often assumes to be all zeros for simplicity. The partial derivatives can then be obtained via Jacobian matrices. Note that besides vanilla LRP, there are other popular variants including LRP-$\epsilon$ and LRP-$\alpha\beta$~\cite{bach2015pixel}. They only differ in the way of scaling $z^{l}$. We use LRP-$\alpha\beta$ for our analysis. As noted in~\citet{pmlr-v70-shrikumar17a}, LRP is equivalent to GI if the neural network only contains linear layers with monotonic non-linear gates (i.e., ReLU). See Appendix~\ref{app:grad-lrp} for explanations.

\subsection{Layerwise Attention Tracing}
Recent works propose token-level Layerwise Attention Tracing (LAT) to track relevance scores $\mathbf{R}_{\text{LAT}}$ of tokens using only attention weights~\cite{abnar-zuidema-2020-quantifying, wu-etal-2020-structured}. Similar to $\mathbf{R}_{\text{LRP}}$, it obeys the conservation law. It starts with an unit relevance score for a sequence and redistribute the score across tokens using self-attention weights by ignoring all other connections. Formally, the redistribution rule is defined as:
\begin{align}
    \mathbf{R}_{\text{LAT}}^{i \leftarrow j} = \sum_{h} \mathbf{A}_{\text{Self-Attn}}^{(h)} \mathbf{R}_{\text{LAT}}^{j(h)} \label{eqn:lat-formula}
\end{align}
where $h$ is the head index, $\mathbf{A}_{\text{Self-Attn}}^{(h)}$ is the learnt \emph{softmax} attention weights~\footnote{\citet{abnar-zuidema-2020-quantifying} also considered residual layers but they are ignored here for simplicity.}.

\begin{figure*}[tp]
\minipage{0.98\textwidth}
  \includegraphics[width=\linewidth]{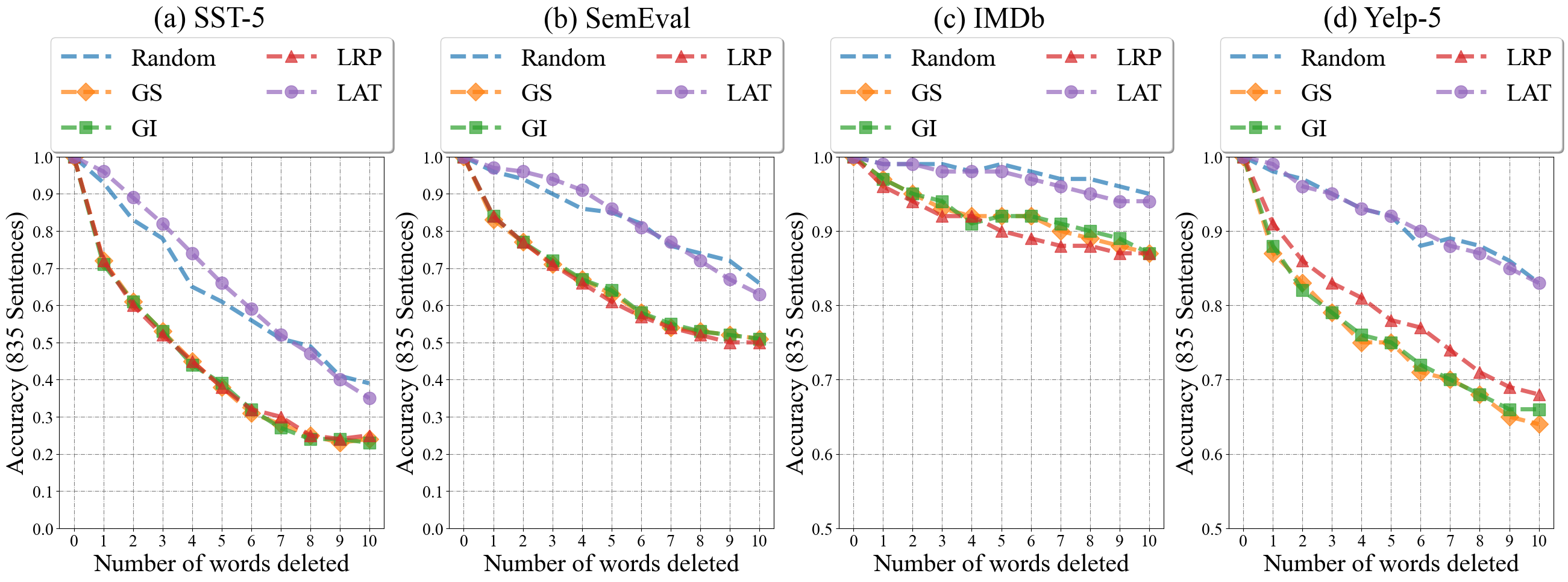}
\endminipage 
\caption{Accuracy v.s. Number of words deleted on initially correctly classified sentences in test sets for all datasets, using attribution methods including Gradient Sensitivity (GS), Gradient $\times$ Input (GI), Layerwise Relevance Propagation (LRP) and Layerwise Attention Tracing (LAT). We also include random word deletion as a baseline.}
\label{fig:word_deletion}
\end{figure*}

\begin{table}[tp]
  \centering
  \setlength{\tabcolsep}{10pt}
  \begin{tabular}[c]{lrrrr}
    \toprule
    Dataset & GS  & GI & LRP & LAT  \\
    \midrule
    SST-5 & 0.91 & 0.91 & 0.89  & 0.85 \\
    SemEval & 0.89 & 0.88 & 0.83  & 0.81 \\
    IMDb & 0.81 & 0.82 & 0.81  & 0.79 \\
    Yelp-5 & 0.83 & 0.86 & 0.79  & 0.73 \\
    \bottomrule
  \end{tabular}
  \caption{Correlations of relevance scores across two models trained with different initializations. }
  \label{tab:random-seed}
\end{table}

\begin{figure}[tb]
  \centering
  \includegraphics[width=0.9\columnwidth]{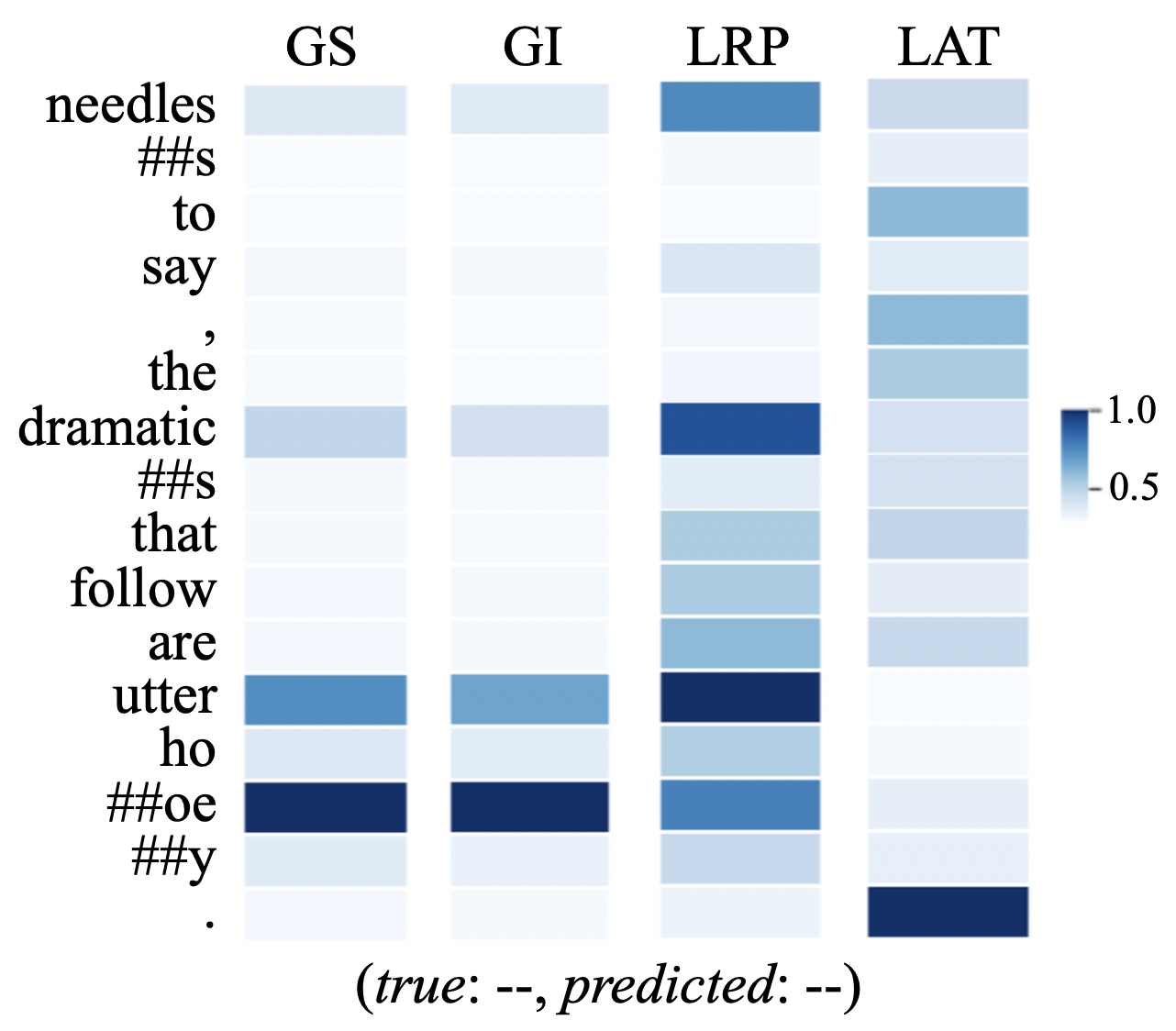}
  \caption{Examples of relevance scores generated using four introduced methods. The model predicts correctly, where the \emph{true} label of this sentence is \emph{very negative} (--).}
  \label{fig:heatmap}
\end{figure}

\section{Models}
We first fine-tuned a BERT model with a classification head for each sentiment analysis dataset from a distinct domain: SST-5 (short sequence movie reviews)~\cite{socher2013recursive}, SemEval~\cite{rosenthal-etal-2017-semeval} (short sequence tweets), IMDb Review~\cite{maas2011learning} (long sequence movie reviews) and Yelp Review (long sequence restaurant reviews)~\cite{zhang2015character}. For additional details on set-ups, see \appref{app:model}.

\section{Experiment}

\begin{figure*}[h]
\minipage{1\textwidth}
  \includegraphics[width=\linewidth]{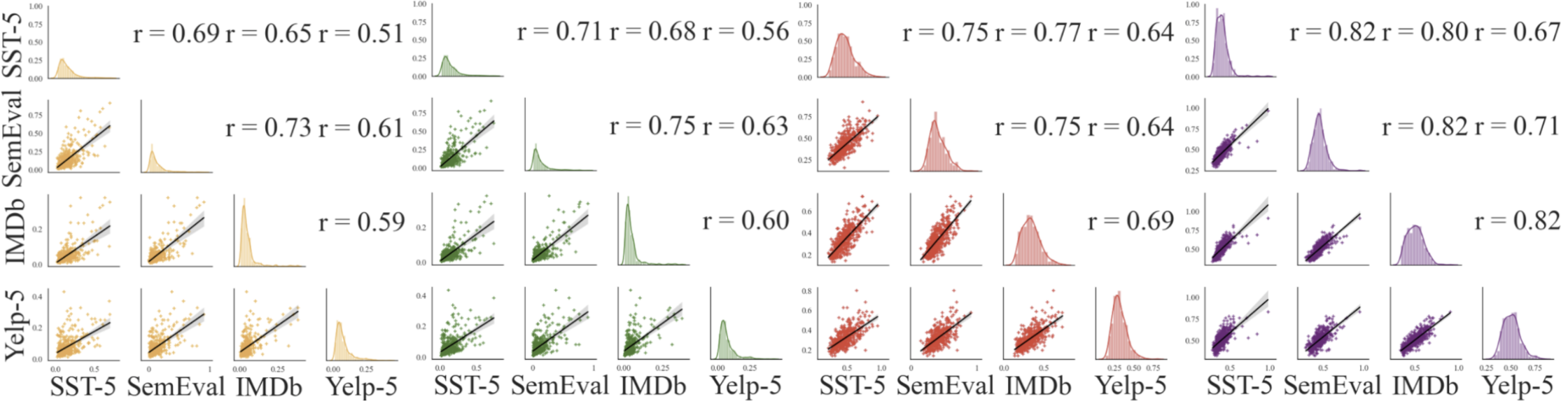}
\endminipage 
\caption{Correlations of relevance scores derived from four attribution methods across four datasets.}
\label{fig:corr_plot}
\end{figure*}

\subsection{Performance and Relevance Scores}
We first used each attribution method to calculate relevance scores tokens of sentences in each test set. Fig.~\ref{fig:heatmap} presents a sentence with token-level relevance scores. Our results suggest that LRP shows richer variance in terms of relevance scores, and encodes richer semantic properties (i.e., all words with strong emotional semantics are highlighted). GS and GI show similar patterns but with more pin-pointed focuses. On the other hand, LAT seems to distract to irrelevant words. \Tabref{tab:top10sst} presents top 5 words and bottom 5 words ranked by their relevance scores for each attribution method. For details on other datasets, see \appref{app:wordrank}.

We further quantified the validity of each attribution methods via an ablation study where we successively remove words in the descending order of relevance scores with in sequence. Fig.~\ref{fig:word_deletion} shows model accuracy drops consistently by removing important words across each dataset. Our results suggest that GS, GI and LRP present similar trends in changes of performance, whereas LAT is close to the trend of random word dropping. This further suggests that LAT may be ineffective in retrieving relevance scores within a sequence.

\begin{table}[tp]
  \centering
  \setlength{\tabcolsep}{4pt}
  \begin{tabular}[c]{llll}
    \toprule
    GS        & GI & LRP & LAT  \\
    \midrule
    mess        & mess        & repetitive  & repetitive \\
    disturbing  & slick       & lacks       & in \\
    slick   & disturbing     & anything   & disturbing \\
    fine      & fine        & hilarious  & lacks \\
    laugh        & miss   & thriller      & stupid \\
    \midrule
    ka    & create     & post       & el \\
    create   & ka      & close          & des \\
    were & were        & el     & lo \\
    ni    & the           & ga      & ka \\
    michael    & michael    & ca       & ho \\
    \bottomrule
  \end{tabular}
  \caption{Top and bottom 5 words ranked by averaged relevance scores derived from each method for SST-5 test. Each words appear at least 5 times. }
  \label{tab:top10sst}
\end{table}

\subsection{Robustness with Random Initialization}
Recent works suggest that fine-tuning is susceptible to random initializations, where model performances may vary significantly. In this vein, we tested whether relevance scores are susceptible with random initializations. We retrained our model with different initialization, and correlate relevance scores. Table~\ref{tab:random-seed} shows Pearson Correlations~\cite{benesty2009pearson} of relevance scores under two different initializations of our fine-tuning process. Our results suggest that random initializations affect our results but only to a limited extend. Furthermore, consistencies in results of longer sentences seem to suffer more from random initializations comparing to shorter ones.

\subsection{ Consistency across Datasets}
We then tested whether attribution methods can explain generalizable semantics through relevance scores across multiple datasets that share similar semantics. We first took a common vocabulary across four datasets, and calculated Pearson Correlations~\cite{benesty2009pearson} of relevance scores for shared words in the vocabulary. Fig.~\ref{fig:corr_plot} presents correlation matrices across four datasets for each attribution methods. Surprisingly, four methods show strong correlations across four datasets, even for LAT which performs poorly in our ablation tests. This suggests that learnt weights in BERT may be generalizable across tasks with similar semantics.

\section{Related Works}
% Left here for later, as the length may vary for other more important parts.
Attribution methods were used to explain BERT in sequence to sequence modelings. For instance, previous works showed that heads can be ranked and pruned via importance scores derived using LRP~\cite{voita-etal-2019-analyzing}. In parallel, recent works used LRP to explain linguistic properties learnt by BERT for machine translation tasks~\cite{voita2020analyzing}. Others applied gradient-based methods and attention weights-based methods to study linguistic properties encoded in self-attention layers~\cite{hao2020self, wu-etal-2020-structured}. Attribution methods are also widely used for explaining other neural networks such as MLP~\cite{li2016visualizing} and LSTM~\cite{arras2017explaining}.

\section{Conclusions}
In this paper, we adapt four existing attribution methods with BERT models for four sequence classification tasks that share similar semantics. Furthermore, we both qualitatively and quantitatively compare the validity and robustness for these four different methods across our datasets. More importantly, we show that attribution methods generalize well across tasks with shared semantics. Our works provide an initial guidance in selecting attribution methods for BERT-based models for downstream probing tests on classification tasks.  

\bibliography{acl2020}

\begin{thebibliography}{30}
\expandafter\ifx\csname natexlab\endcsname\relax\def\natexlab#1{#1}\fi

\bibitem[{Abnar and Zuidema(2020)}]{abnar-zuidema-2020-quantifying}
Samira Abnar and Willem Zuidema. 2020.
\newblock \href {https://doi.org/10.18653/v1/2020.acl-main.385} {Quantifying
  attention flow in transformers}.
\newblock In \emph{Proceedings of the 58th Annual Meeting of the Association
  for Computational Linguistics}, pages 4190--4197, Online. Association for
  Computational Linguistics.

\bibitem[{Arras et~al.(2017)Arras, Montavon, M{\"u}ller, and
  Samek}]{arras2017explaining}
Leila Arras, Gr{\'e}goire Montavon, Klaus-Robert M{\"u}ller, and Wojciech
  Samek. 2017.
\newblock Explaining recurrent neural network predictions in sentiment
  analysis.
\newblock In \emph{Proceedings of the 8th Workshop on Computational Approaches
  to Subjectivity, Sentiment and Social Media Analysis}, pages 159--168.

\bibitem[{Bach et~al.(2015)Bach, Binder, Montavon, Klauschen, M{\"u}ller, and
  Samek}]{bach2015pixel}
Sebastian Bach, Alexander Binder, Gr{\'e}goire Montavon, Frederick Klauschen,
  Klaus-Robert M{\"u}ller, and Wojciech Samek. 2015.
\newblock On pixel-wise explanations for non-linear classifier decisions by
  layer-wise relevance propagation.
\newblock \emph{PloS one}, 10(7).

\bibitem[{Benesty et~al.(2009)Benesty, Chen, Huang, and
  Cohen}]{benesty2009pearson}
Jacob Benesty, Jingdong Chen, Yiteng Huang, and Israel Cohen. 2009.
\newblock Pearson correlation coefficient.
\newblock In \emph{Noise reduction in speech processing}, pages 1--4. Springer.

\bibitem[{B{\"o}hle et~al.(2019)B{\"o}hle, Eitel, Weygandt, and
  Ritter}]{bohle2019layer}
Moritz B{\"o}hle, Fabian Eitel, Martin Weygandt, and Kerstin Ritter. 2019.
\newblock Layer-wise relevance propagation for explaining deep neural network
  decisions in mri-based alzheimer’s disease classification.
\newblock \emph{Frontiers in aging neuroscience}, 11:194.

\bibitem[{Clark et~al.(2019)Clark, Khandelwal, Levy, and
  Manning}]{clark2019what}
Kevin Clark, Urvashi Khandelwal, Omer Levy, and Christopher~D Manning. 2019.
\newblock What does {BERT} look at? {A}n analysis of {BERT}'s attention.
\newblock In \emph{Proceedings of the 2019 ACL Workshop BlackBoxNLP: Analyzing
  and Interpreting Neural Networks for NLP}, pages 276--286.

\bibitem[{Clark et~al.(2020)Clark, Luong, Le, and Manning}]{clark2020electra}
Kevin Clark, Minh-Thang Luong, Quoc~V Le, and Christopher~D Manning. 2020.
\newblock Electra: Pre-training text encoders as discriminators rather than
  generators.
\newblock \emph{arXiv preprint arXiv:2003.10555}.

\bibitem[{Devlin et~al.(2019)Devlin, Chang, Lee, and
  Toutanova}]{devlin2019bert}
Jacob Devlin, Ming-Wei Chang, Kenton Lee, and Kristina Toutanova. 2019.
\newblock {BERT}: Pre-training of deep bidirectional transformers for language
  understanding.
\newblock In \emph{Proceedings of the 2019 Conference of the North American
  Chapter of the Association for Computational Linguistics: Human Language
  Technologies}.

\bibitem[{Goldberg(2019)}]{goldberg2019assessing}
Yoav Goldberg. 2019.
\newblock Assessing {BERT}'s syntactic abilities.
\newblock \emph{arXiv preprint arXiv:1901.05287}.

\bibitem[{Hao et~al.(2020)Hao, Dong, Wei, and Xu}]{hao2020self}
Yaru Hao, Li~Dong, Furu Wei, and Ke~Xu. 2020.
\newblock Self-attention attribution: Interpreting information interactions
  inside transformer.
\newblock \emph{arXiv preprint arXiv:2004.11207}.

\bibitem[{Hendrycks and Gimpel(2016)}]{hendrycks2016gaussian}
Dan Hendrycks and Kevin Gimpel. 2016.
\newblock Gaussian error linear units (gelus).
\newblock \emph{arXiv preprint arXiv:1606.08415}.

\bibitem[{Hewitt and Manning(2019)}]{hewitt2019structural}
John Hewitt and Christopher~D Manning. 2019.
\newblock A structural probe for finding syntax in word representations.
\newblock In \emph{Proceedings of the 2019 Conference of the North American
  Chapter of the Association for Computational Linguistics: Human Language
  Technologies, Volume 1 (Long and Short Papers)}, pages 4129--4138.

\bibitem[{Kindermans et~al.(2017)Kindermans, Hooker, Adebayo, Alber,
  Sch{\"u}tt, D{\"a}hne, Erhan, and Kim}]{kindermans2017reliability}
Pieter-Jan Kindermans, Sara Hooker, Julius Adebayo, Maximilian Alber, Kristof~T
  Sch{\"u}tt, Sven D{\"a}hne, Dumitru Erhan, and Been Kim. 2017.
\newblock The (un) reliability of saliency methods.
\newblock \emph{arXiv preprint arXiv:1711.00867}.

\bibitem[{Li et~al.(2016)Li, Chen, Hovy, and Jurafsky}]{li2016visualizing}
Jiwei Li, Xinlei Chen, Eduard Hovy, and Dan Jurafsky. 2016.
\newblock Visualizing and understanding neural models in {NLP}.
\newblock In \emph{Proceedings of the 2016 Conference of the North American
  Chapter of the Association for Computational Linguistics: Human Language
  Technologies}.

\bibitem[{Liu et~al.(2019)Liu, Ott, Goyal, Du, Joshi, Chen, Levy, Lewis,
  Zettlemoyer, and Stoyanov}]{liu2019roberta}
Yinhan Liu, Myle Ott, Naman Goyal, Jingfei Du, Mandar Joshi, Danqi Chen, Omer
  Levy, Mike Lewis, Luke Zettlemoyer, and Veselin Stoyanov. 2019.
\newblock Roberta: A robustly optimized bert pretraining approach.
\newblock \emph{arXiv preprint arXiv:1907.11692}.

\bibitem[{Maas et~al.(2011)Maas, Daly, Pham, Huang, Ng, and
  Potts}]{maas2011learning}
Andrew Maas, Raymond~E Daly, Peter~T Pham, Dan Huang, Andrew~Y Ng, and
  Christopher Potts. 2011.
\newblock Learning word vectors for sentiment analysis.
\newblock In \emph{Proceedings of the 49th annual meeting of the association
  for computational linguistics: Human language technologies}, pages 142--150.

\bibitem[{Raganato and Tiedemann(2018)}]{raganato2018analysis}
Alessandro Raganato and J{\"o}rg Tiedemann. 2018.
\newblock An analysis of encoder representations in {T}ransformer-based machine
  translation.
\newblock In \emph{Proceedings of the 2018 EMNLP Workshop BlackboxNLP:
  Analyzing and Interpreting Neural Networks for NLP}, pages 287--297.

\bibitem[{Rosenthal et~al.(2017)Rosenthal, Farra, and
  Nakov}]{rosenthal-etal-2017-semeval}
Sara Rosenthal, Noura Farra, and Preslav Nakov. 2017.
\newblock \href {https://doi.org/10.18653/v1/S17-2088} {{S}em{E}val-2017 task
  4: Sentiment analysis in {T}witter}.
\newblock In \emph{Proceedings of the 11th International Workshop on Semantic
  Evaluation ({S}em{E}val-2017)}, pages 502--518, Vancouver, Canada.
  Association for Computational Linguistics.

\bibitem[{Shrikumar et~al.(2017)Shrikumar, Greenside, and
  Kundaje}]{pmlr-v70-shrikumar17a}
Avanti Shrikumar, Peyton Greenside, and Anshul Kundaje. 2017.
\newblock \href {http://proceedings.mlr.press/v70/shrikumar17a.html} {Learning
  important features through propagating activation differences}.
\newblock In \emph{Proceedings of the 34th International Conference on Machine
  Learning}, volume~70 of \emph{Proceedings of Machine Learning Research},
  pages 3145--3153, International Convention Centre, Sydney, Australia. PMLR.

\bibitem[{Socher et~al.(2013)Socher, Perelygin, Wu, Chuang, Manning, Ng, and
  Potts}]{socher2013recursive}
Richard Socher, Alex Perelygin, Jean Wu, Jason Chuang, Christopher~D Manning,
  Andrew~Y Ng, and Christopher Potts. 2013.
\newblock Recursive deep models for semantic compositionality over a sentiment
  treebank.
\newblock In \emph{Proceedings of the 2013 Conference on Empirical Methods in
  Natural Language Processing}, pages 1631--1642.

\bibitem[{Tenney et~al.(2019)Tenney, Das, and Pavlick}]{tenney2019bert}
Ian Tenney, Dipanjan Das, and Ellie Pavlick. 2019.
\newblock Bert rediscovers the classical nlp pipeline.
\newblock In \emph{Proceedings of the 57th Annual Meeting of the Association
  for Computational Linguistics}, pages 4593--4601.

\bibitem[{Vig and Belinkov(2019)}]{vig2019analyzing}
Jesse Vig and Yonatan Belinkov. 2019.
\newblock Analyzing the structure of attention in a transformer language model.
\newblock In \emph{Proceedings of the 2019 ACL Workshop BlackboxNLP: Analyzing
  and Interpreting Neural Networks for NLP}, pages 63--76.

\bibitem[{Voita et~al.(2020)Voita, Sennrich, and Titov}]{voita2020analyzing}
Elena Voita, Rico Sennrich, and Ivan Titov. 2020.
\newblock Analyzing the source and target contributions to predictions in
  neural machine translation.
\newblock \emph{arXiv preprint arXiv:2010.10907}.

\bibitem[{Voita et~al.(2018)Voita, Serdyukov, Sennrich, and
  Titov}]{voita2018context}
Elena Voita, Pavel Serdyukov, Rico Sennrich, and Ivan Titov. 2018.
\newblock Context-aware neural machine translation learns anaphora resolution.
\newblock In \emph{Proceedings of the 56th Annual Meeting of the Association
  for Computational Linguistics}, pages 1264--1274.

\bibitem[{Voita et~al.(2019)Voita, Talbot, Moiseev, Sennrich, and
  Titov}]{voita-etal-2019-analyzing}
Elena Voita, David Talbot, Fedor Moiseev, Rico Sennrich, and Ivan Titov. 2019.
\newblock \href {https://doi.org/10.18653/v1/P19-1580} {Analyzing multi-head
  self-attention: Specialized heads do the heavy lifting, the rest can be
  pruned}.
\newblock In \emph{Proceedings of the 57th Annual Meeting of the Association
  for Computational Linguistics}, pages 5797--5808, Florence, Italy.
  Association for Computational Linguistics.

\bibitem[{Wang et~al.(2018)Wang, Singh, Michael, Hill, Levy, and
  Bowman}]{wang2018glue}
Alex Wang, Amanpreet Singh, Julian Michael, Felix Hill, Omer Levy, and Samuel
  Bowman. 2018.
\newblock Glue: A multi-task benchmark and analysis platform for natural
  language understanding.
\newblock In \emph{Proceedings of the 2018 EMNLP Workshop BlackboxNLP:
  Analyzing and Interpreting Neural Networks for NLP}, pages 353--355.

\bibitem[{Wu et~al.(2020)Wu, Nguyen, and Ong}]{wu-etal-2020-structured}
Zhengxuan Wu, Thanh-Son Nguyen, and Desmond Ong. 2020.
\newblock \href {https://doi.org/10.18653/v1/2020.blackboxnlp-1.24} {Structured
  self-{A}ttention{W}eights encode semantics in sentiment analysis}.
\newblock In \emph{Proceedings of the Third BlackboxNLP Workshop on Analyzing
  and Interpreting Neural Networks for NLP}, pages 255--264, Online.
  Association for Computational Linguistics.

\bibitem[{Yang et~al.(2018)Yang, Tresp, Wunderle, and
  Fasching}]{yang2018explaining}
Yinchong Yang, Volker Tresp, Marius Wunderle, and Peter~A Fasching. 2018.
\newblock Explaining therapy predictions with layer-wise relevance propagation
  in neural networks.
\newblock In \emph{2018 IEEE International Conference on Healthcare Informatics
  (ICHI)}, pages 152--162. IEEE.

\bibitem[{Yang et~al.(2019)Yang, Dai, Yang, Carbonell, Salakhutdinov, and
  Le}]{yang2019xlnet}
Zhilin Yang, Zihang Dai, Yiming Yang, Jaime Carbonell, Russ~R Salakhutdinov,
  and Quoc~V Le. 2019.
\newblock Xlnet: Generalized autoregressive pretraining for language
  understanding.
\newblock In \emph{Advances in neural information processing systems}, pages
  5753--5763.

\bibitem[{Zhang et~al.(2015)Zhang, Zhao, and LeCun}]{zhang2015character}
Xiang Zhang, Junbo Zhao, and Yann LeCun. 2015.
\newblock Character-level convolutional networks for text classification.
\newblock \emph{Advances in neural information processing systems},
  28:649--657.

\end{thebibliography}
\bibliographystyle{acl_natbib}

\newpage
\clearpage

\appendix

\section*{Appendix}

\section{Non-linear Activations}\label{app:non-linear}
Point-wise monotonic activation functions are ignored by LRP as the estimated gradients for these layer are not changed for a given $x$. We are aware that the GeLU activations~\cite{hendrycks2016gaussian} in BERT is not strictly monotonically increasing, and we leave this for the future to investigate potential drawbacks of ignoring this in BERT. We split the test set evenly if the original dataset does not come with a dev set.

\section{Gradient-based methods and LRP}\label{app:grad-lrp}
In this section, we present connections between gradient-based attribution methods (e.g., GS or GI) and LRP. First, let us rewrite our Eqn.~\ref{eqn:uni-gradient} using chain-rule as~\cite{bach2015pixel}:
\begin{align}
    \mathbf{R}^{\text{GS}}_{i} (x) =  \frac{\partial f_{c}(x)}{\partial x_{i}^{l}} \frac{\partial x_{i}^{l}}{\partial x_{i}^{l-1}} \cdots \frac{\partial x_{i}^{1}}{\partial x_{i}} \label{eqn:uni-gradient-chain-rule}
\end{align}
where $x_{i}^{l}$ are intermediate hidden states in different layers. Let us consider the simplest case where the neural network contains only linear layers with non-linear activation functions $g(\cdot)$ as in Eqn.~\ref{eqn:linear-layer-act-1} to ~\ref{eqn:linear-layer-act-2}. We can then rewrite Eqn.~\ref{eqn:uni-gradient-chain-rule} as:
\begin{align}
    \mathbf{R}^{\text{GS}}_{i} (x) &= \mathbf{w}^{l} g^{\prime}(z_{j}^{l+1}) \cdots \mathbf{w}^{0} g^{\prime}(z_{j}^{1})  \label{eqn:uni-gradient-chain-rule-1} \\
    &= (\prod_{l} \mathbf{w}^{l}) (\prod_{l} g^{\prime}(z_{j}^{l+1}))
    \label{eqn:uni-gradient-chain-rule-2}
\end{align}
where $g^{\prime}(\cdot)$ is the derivatives of non-linear activation functions. For GI, we need to append the input features to the front as in Eqn.~\ref{eqn:uni-gradient-input}:
\begin{align}
    \mathbf{R}^{\text{GI}}_{i} (x) &= x_{i} (\prod_{l} \mathbf{w}^{l}) (\prod_{l} g^{\prime}(z_{j}^{l+1}))
    \label{eqn:uni-gradient-chain-rule-2}
\end{align}
To reiterate our formulation for LRP as in Eqn.~\ref{eqn:lrp-general-1}:
\begin{align}
    \mathbf{R}^{\text{LRP}}_{i} (x) &= f_{c}(x) (\prod_{l} \frac{\mathbf{w}^{l} x^{l}}{z^{l+1}}) (\prod_{l} g^{\prime}(\mathbf{z}_{j}^{l+1}))
\end{align}
where, if we ignore non-linear activation functions, \citet{pmlr-v70-shrikumar17a} showed that absent modifications for numerical stability, vanilla LRP rules were equivalent within a scaling factor of GS. See \citet{pmlr-v70-shrikumar17a} for a comprehensive proof.

\section{Datasets and Models}\label{app:model}

We fine-tuned BERT with four datasets separately. \Tabref{tab:datasets} presents statistics of our datasets. \Tabref{tab:datasets} shows model performance for each dataset. Our results show that all of our models achieved the state-of-the-art performance~\cite{wang2018glue}.

Our fine-tune begins with the uncased BERT-base parameters~\footnote{\url{https://storage.googleapis.com/bert_models/2020_02_20/uncased_L-12_H-768_A-12.zip}} and adds a n-way sentiment classifier head. During fine-tuning, BERT-base is trained for 3 epochs where the best model is recorded. As in the original BERT-base model \citep{liu2019roberta}, our model consists 12 heads and 12 layers, with hidden layer size 768. The model uses the default BERT WordPiece tokenizer, with a maximum sequence length of 512. The initial learning rate is $2e^{-5}$ for all trainable parameters, with a batch size of 8 per device (i.e., GPU). We fine-tuned for 3 epochs with a dropout probability of 0.1 for both attention weights and hidden states. The Best model is chosen based on performance on the respective dev set.

We used 6 $\times$ GeForce RTX 2080 Ti GPU each with 11GB memory to fine-tune. The fine-tuning process takes from 1 hour to 10 hours to finish from the smallest dataset to the largest one.

\begin{table}[hp]
  \centering
  \setlength{\tabcolsep}{5pt}
  \begin{tabular}[c]{l *{4}{r}}
    \toprule
    Dataset        & Train & Dev & Test & Acc /\% \\
    \midrule
    SST-5            & 156,817    & 1,101    & 2,210  & 56.9 \\
    SemEval          & 39,656     & 2,478    & 2,478  & 74.2 \\
    IMDb             & 24,999     & 12,500   & 12,499 & 89.0 \\
    Yelp-5           & 649,999    & 25,000   & 24,999 & 66.0 \\
    \bottomrule
  \end{tabular}
  \caption{Datasets with model performances.}
  \label{tab:datasets}
\end{table}

\section{Word Rankings}\label{app:wordrank}
These tokens are from randomly sampled 2000 sentences from SST-5, and at least appear five times. The reason to filter words based on frequency is that we want to have more stable results by taking the average relevance score across many occurrences. We also exclude punctuation marks in the \Tabref{tab:top10sst} by focusing on words which are linguistically interesting. Table~\ref{tab:top10semeval} to ~\ref{tab:top10yelp} provide tables with top 5 and bottom 5 words ranked by relevance scores for other three datasets for comparison.

\begin{table*}[tp]
  \centering
  \setlength{\tabcolsep}{4pt}
  \begin{tabular}[c]{llll}
    \toprule
    GS        & GI & LRP & LAT  \\
    \midrule
    weird        & weird        & excited  & in \\
    sad        & nice        & awesome       & undertaker \\
    nice   & sad            & jurassic   & to \\
    sweet     & sweet        & amazing  & fleetwood \\
    excellent        & excellent   & happy      & excited \\
    \midrule
    co    & 29     & green       & sy \\
    29   & co      & return          & wo \\
    30 & 30        & stream     & mi \\
    48    & 48       & fr      & je \\
    ar    & 21    & en       & ac \\
    \bottomrule
  \end{tabular}
  \caption{Top 5 words followed by bottom 5 words with ranked by averaged relevance scores based on different attribution methods for SemEval test. Each words appear at least 5 times. Punctuation marks are skipped. }
  \label{tab:top10semeval}
\end{table*}

\begin{table*}[tp]
  \centering
  \setlength{\tabcolsep}{4pt}
  \begin{tabular}[c]{llll}
    \toprule
    GS        & GI & LRP & LAT  \\
    \midrule
    succeeds        & succeeds        & simpsons            & mtv \\
    nightmare        & perfection        & emotionally       & countless \\
    joke        & nightmare      & mtv                    & emotionally \\
    creep      & joke              & animated            & everyday \\
    perfection        & creep   & immensely               & include \\
    \midrule
    passed    & passed     & fist       & br \\
    rave   & rose      & semi          & mon \\
    rose & ernest     & stu    & ac \\
    ernest    & rave        & mon      & mar \\
    pac    & feet    & cad       & hal \\
    \bottomrule
  \end{tabular}
  \caption{Top 5 words followed by bottom 5 words with ranked by averaged relevance scores based on different attribution methods for IMDb test. Each words appear at least 5 times. Punctuation marks are skipped. }
  \label{tab:top10imdb}
\end{table*}

\begin{table*}[tp]
  \centering
  \setlength{\tabcolsep}{4pt}
  \begin{tabular}[c]{llll}
    \toprule
    GS        & GI & LRP & LAT  \\
    \midrule
    tooth        & tooth        & waitress  & email \\
    edible        & edible       & amazing       & amazing \\
    superb   & superb     & disappointing   & dessert \\
    thrilled      & grade        & detail  & convenience \\
    pink        & odd   & disgusting      & upset \\
    \midrule
    tar    & tar    & tar       & lu \\
    tree   & tree      & rico          & ha \\
    lu & lu                    & nr     & ni \\
    keeping    & national        & trees      & z \\
    ke    & keeping           & nun       & ke \\
    \bottomrule
  \end{tabular}
  \caption{Top 5 words followed by bottom 5 words with ranked by averaged relevance scores based on different attribution methods for Yelp-5 test. Each words appear at least 5 times. Punctuation marks are skipped. }
  \label{tab:top10yelp}
\end{table*}

\end{document}